\documentclass[letterpaper, 10 pt, conference]{ieeeconf}
\IEEEoverridecommandlockouts
\usepackage{cite}
\usepackage{amsmath,amssymb,amsfonts}
\usepackage{algorithmic}
\usepackage{graphicx}
\usepackage{textcomp}
\usepackage{xcolor}
\usepackage{booktabs}
\usepackage{tabularx}
\usepackage{url}
\usepackage{xcolor}
\usepackage{todonotes}
\usepackage{tikz}
\usepackage{float}
\def\BibTeX{{\rm B\kern-.05em{\sc i\kern-.025em b}\kern-.08em
    T\kern-.1667em\lower.7ex\hbox{E}\kern-.125emX}}
\begin{document}

\title{A Bayesian Reasoning Framework \\for Robotic Systems in Autonomous Casualty Triage
}

\author{Szymon Rusiecki$^{1, 2}$, Cecilia Morales$^{2}$, Pia Störy$^{2, 3}$, \\ Kimberly Elenberg$^{2}$, Leonard Weiss$^{4}$, and Artur Dubrawski$^{2}$%
\thanks{$^{1}$AGH University of Krakow, Kraków, Poland\newline{\tt\small rusiecki@student.agh.edu.pl}}%
\thanks{$^{2}$Carnegie Mellon University, Pittsburgh, PA, USA\newline{\tt\small \{cgmorale, kelenber, awd\}@andrew.cmu.edu}}%
\thanks{$^{3}$Osnabrück University, Osnabrück, Germany\newline{\tt\small pstoery@uni-osnabrueck.de}}%
\thanks{$^{4}$University of Pittsburgh, Pittsburgh, PA, USA\newline{\tt\small weissls2@upmc.edu}}%
}

\newcommand\copyrighttext{%
  \footnotesize © 2026 IEEE. Personal use of this material is permitted. Permission from IEEE must be obtained for all other uses, in any current or future media, including reprinting/republishing this material for advertising or promotional purposes, creating new collective works, for resale or redistribution to servers or lists, or reuse of any copyrighted component of this work in other works.}
\newcommand\copyrightnotice{%
\begin{tikzpicture}[remember picture,overlay]
\node[anchor=south,yshift=10pt] at (current page.south) {\fbox{\parbox{\dimexpr\textwidth-\fboxsep-\fboxrule\relax}{\copyrighttext}}};
\end{tikzpicture}%
}

\maketitle
\copyrightnotice
\thispagestyle{empty}
\pagestyle{empty}

\begin{abstract}
Autonomous robots deployed in mass casualty incidents (MCI) face the challenge of making critical decisions based on incomplete and noisy perceptual data. We present an autonomous robotic system for casualty assessment that fuses outputs from multiple vision-based algorithms, estimating signs of severe hemorrhage, visible trauma, or physical alertness, into a coherent triage assessment. At the core of our system is a Bayesian network, constructed from expert-defined rules, which enables probabilistic reasoning about a casualty's condition even with missing or conflicting sensory inputs. The system,  evaluated during the DARPA Triage Challenge (DTC) in realistic MCI scenarios  involving 11 and 9 casualties, demonstrated a nearly three-fold improvement in physiological assessment accuracy (from 15\% to 42\% and 19\% to 46\%) compared to a vision-only baseline. More importantly, overall triage accuracy increased from 14\% to 53\%, while the diagnostic coverage of the system expanded from 31\% to 95\% of cases. These results demonstrate that integrating expert-guided probabilistic reasoning with advanced vision-based sensing can significantly enhance the reliability and decision-making capabilities of autonomous systems in critical real-world applications.
\end{abstract}

\section{Introduction}
\label{sec:intro}

When a terrorist attack struck the Boston Marathon in 2013, emergency responders faced around 250 casualties in minutes, overwhelming the city's trauma treatment capacity and forcing life-or-death triage decisions under extreme pressure~\cite{gunaratna2013current}. Such mass casualty incidents (MCIs), resulting from terrorist attacks, armed conflicts, or large-scale natural disasters, can involve hundreds of victims and instantly overwhelm local emergency services~\cite{tahernejad2024application}. In these chaotic scenarios, conventional manual triage methods can fail when faced with high volumes of casualties and complex injury profiles~\cite{super1994start, gabbe2022review}.

To address these challenges, autonomous robotic systems offer a compelling solution, capable of surveying disaster zones to identify victims and remotely assess their injuries~\cite{hashimoto2017four, agarwal2014characteristics, adam2023using}. However, deploying these robots in real-world MCI environments introduces a unique set of severe operational challenges~\cite{tiwari2019unified, patel2023dream, park2017disaster, wadden2022defining}. Robots must navigate cluttered and unpredictable disaster zones, operate under poor visibility from smoke or dust, and contend with severe sensor occlusions where casualties are partially or fully obscured by debris. Consequently, the data they collect with their sensors are often noisy, incomplete, and fragmented, making a reliable assessment difficult.

\begin{figure}[t]
    \centering
    \includegraphics[width=.8\linewidth]{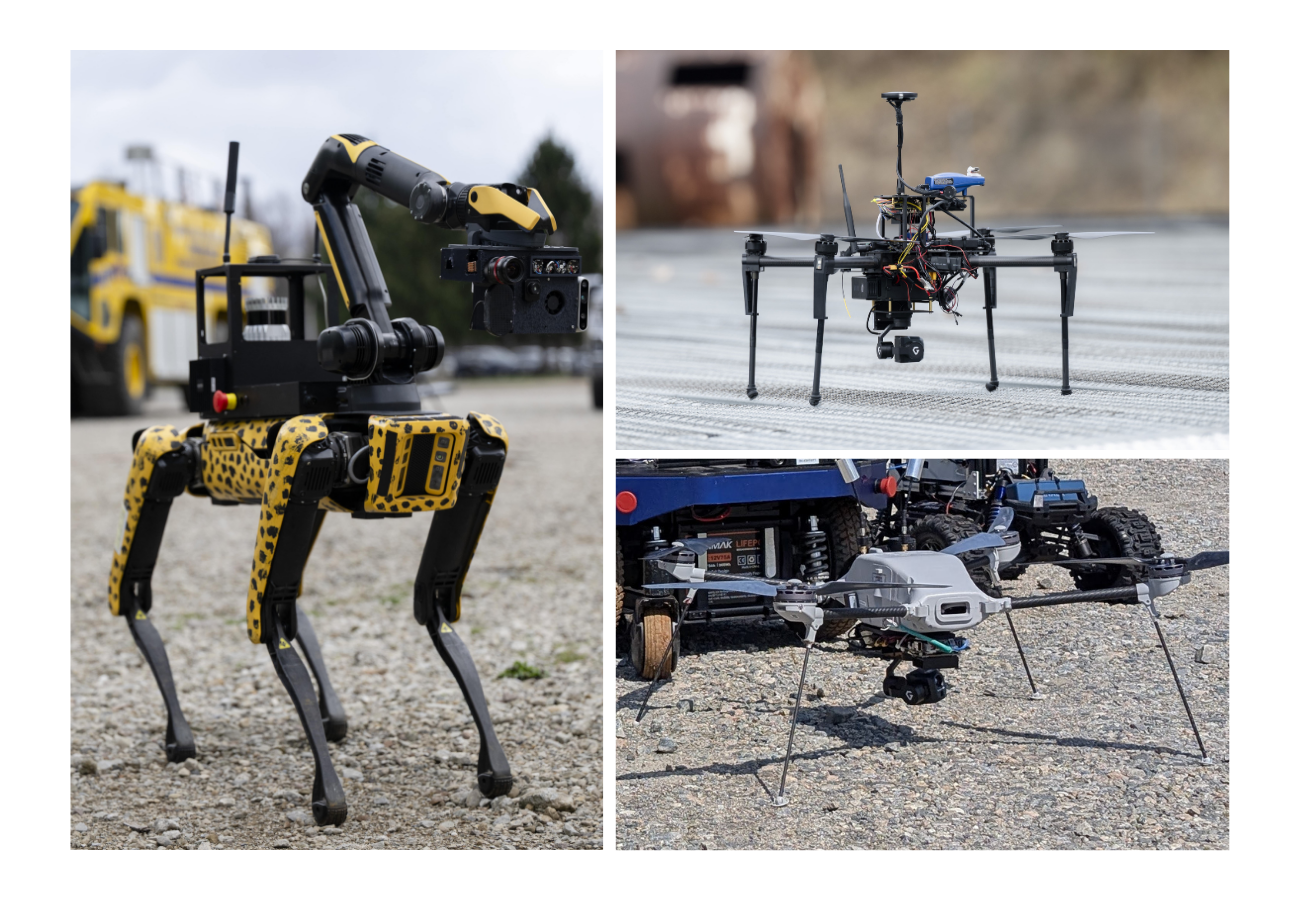}
    \caption{UGV and UAV robots used in our experiments}
    \label{fig:robots}
\end{figure}

This inherent unreliability of sensory data poses a fundamental challenge to the perception modules running on these robots. Current perception algorithms often suffer from three critical limitations in this context \textbf{algorithmic isolation}: treating each vital sign independently, \textbf{context blindness}: ignoring clinical inter-dependencies, and \textbf{brittleness}: underperforming under noisy conditions.

For example, a robot may flag an amputation and a blood detector may see blood, yet without context-aware reasoning the system may not be able to infer that the casualty is actively hemorrhaging, a higher triage priority than an isolated lower-extremity wound. Therefore, autonomous triage requires more than a set of independent detectors; it needs a cognitive architecture that can reason under uncertainty and fuse partial evidence into a coherent assessment.

To overcome these limitations and enable robust robotic decision-making, we propose a cognitive architecture centered on a Bayesian Network (BN). This reasoning engine integrates outputs from multiple perception algorithms into a unified probabilistic estimate of the patient's condition. Constructed from structured clinical knowledge elicited from experts, our BN is designed to support inference even when a sensor fails or a casualty is occluded, improving the consistency and interpretability of the robot's triage decisions in real time. This expert-knowledge foundation provides a robust and transparent alternative for decision support in the data-scarce scenarios typical for robotics in MCIs.

In this work, we introduce a knowledge-based cognitive architecture for automated robotic triage. Our key contributions are (1) a novel architecture for autonomous robotic triage centered on an expert-elicited Bayesian Network that integrates multimodal perception inputs, (2) a robust reasoning framework that enables the robot to handle the noisy, uncertain, and incomplete sensory data characteristic of real-world disaster environments, and (3) a full system validation through realistic field scenarios, demonstrating significantly improved triage accuracy over the alternative employing independent algorithmic outputs. The development and evaluation of this system were carried out in the context of the DARPA Triage Challenge (DTC), an initiative aimed at advancing autonomous medical triage capabilities for mass casualty incidents.

To the best of our knowledge, the proposed system represents one of the first neuro-symbolic frameworks for this application to be validated at scale. Its performance was benchmarked in a comprehensive field evaluation involving 20 distinct casualty cases, demonstrating its robustness outside of a controlled laboratory environment.

\section{Related work}
\label{sec:related_work}

\subsection{Bayesian Networks}
Bayesian Networks (BNs)~\cite{pearl2014probabilistic}, introduced in the 1980s, are Probabilistic Graphical Models that represent uncertain knowledge using Directed Acyclic Graphs (DAGs), where nodes denote random variables and edges represent conditional dependencies. The power of BNs is based on Bayes' Theorem~\cite{bayes1763lii} enabling a compact representation of a joint probability distribution over a set of random variables $\mathcal{X} = \{X_1, \dots, X_n\}$. The structure of the DAG encodes conditional independence assumptions, allowing for a computationally tractable factorization of the joint distribution:
\begin{equation}
\label{eq:pred}
    P(X_1, \dots, X_n) = \prod_{i=1}^{n} P(X_i | \text{Pa}(X_i))
\end{equation}
where $\text{Pa}(X_i)$ denotes the set of parents of variable $X_i$ in the DAG. This factorization significantly reduces the number of parameters required to define the full joint distribution, making even complex probabilistic models feasible.

\subsection{Bayesian Networks in Medical Applications}
BNs have been effectively used for clinical triage by modeling relationships between patient symptoms and outcomes. For instance, BN-based systems have been developed for triaging non-traumatic abdominal pain~\cite{sadeghi2006bayesian} and detecting pediatric asthma with high accuracy~\cite{sanders2006prospective}. While other studies also explore BNs for triage support, they often highlight the complexity of this approach~\cite{abad2008evolution, fareh2023probabilistic, olszewski2003bayesian}.

Concurrently, recent advances in computer vision and machine learning have enabled remote, non-contact prediction of vital signs, such as heart rate from eye movements~\cite{zheng2022heart}, SpO2 levels~\cite{al2021non}, and blood pressure via rPPG~\cite{schrumpf2021assessment}. A recent systematic review and meta-analysis confirms the potential of this approach, showing that data-driven AI models generally outperform conventional triage tools in predicting trauma outcomes~\cite{yuba2022systematic, adebayo2023exploring, vantu2023medical, defilippo2023computational}. However, a significant challenge remains to translate the success of these individual models into a robust integrated system capable of assessing each casualty case as a whole. Relying solely on the isolated predictions of these diverse, often "black-box" neural networks can lead to inaccuracies, particularly in complex environments.

\subsection{Robotics in Search and Rescue}
Depending on the context, robotic systems for triage can be broadly divided into remote and contact solutions. Contact approaches, conventionally used in hospitals, such as wearable monitoring devices~\cite{polley2021wearable}, have a clear limitation: the injured individual must be equipped with the device beforehand, which is often unrealistic in emergency and mass casualty scenarios. This limitation motivates the use of remote solutions. Some approaches~\cite{alvarez2021development, lu2023unmanned} use Unmanned Aerial Vehicles (UAVs) to assess casualties from the air, enabling rapid coverage of large or hard-to-reach scenes. However, short flight times, payload limits, and difficulty maintaining a steady viewpoint in wind, precipitation, or clutter make stable data collection challenging. This is especially problematic for video methods that depend on subtle cues, e.g. detecting respiratory distress via abnormal chest motion, because camera or body movement can mask the thoracic signal; such algorithms typically require stationary, stabilized observations to perform reliably.

On the other hand, Unmanned Ground Vehicles (UGVs), allow closer interaction with victims and the use of more sophisticated sensors. However, UGVs face challenges related to terrain accessibility, navigation in debris-filled areas, and the time required to physically approach each casualty.

These limitations highlight that while both UAV and UGV based approaches expand the possibilities of remote triage, neither solution alone fully addresses the complexities of real-world mass casualty incidents. It is important to emphasize that this field is still in its infancy. Current solutions typically focus on either UAVs or UGVs, each addressing only part of the problem. Equally important is the challenge of multimodal data fusion: learning how to aggregate heterogeneous information streams from visual, thermal, acoustic, and physiological sensors into a coherent and reliable triage decision framework. Addressing these challenges will be key to advancing robotic triage systems from experimental prototypes to effective tools useful in real-world incidents.
\section{Methods}
\label{sec:methodology}

\subsection{Environment}
Our study was conducted in field tests that simulate MCIs with 20 casualties, constructed for the DTC. Experiments were conducted in accordance with the Institutional Review Board. Autonomous ground and aerial robots used only non-contact, stand-off sensors to locate and assess casualties, estimating vital physiological parameters and visible life-threatening indicators to support early triage. The environment featured unstructured terrain, constrained passages, occlusions, and sensory degradation (fog, dust, background noise); some casualties were partially hidden under debris or inside vehicles, requiring active search. Casualties were human volunteers and TOMManikin~\cite{tommanikin} high-fidelity manikins, distributed with varied injuries and postures. Our system fused outputs from multiple onboard perception algorithms within a probabilistic reasoning framework to infer physiological state and injury severity from incomplete observations.

\begin{table}[t]
\centering
\caption{Scoring criteria for emergency scenarios.}
\label{scoring}
\renewcommand{\arraystretch}{1} 
\begin{tabular}{l l l}
\toprule
\textbf{Field} & \textbf{Values} & \textbf{Scoring Criteria} \\
\midrule\midrule
Severe    & Present / Absent & \textbf{4} if in GW \\ Hemorrhage & & \textbf{2} if match GT \\ & & \textbf{0} otherwise \\
\midrule\midrule
Respiratory  & Present / Absent & \textbf{4} if in GW \\ Distress &  & \textbf{2} if match GT \\ & & \textbf{0} otherwise \\
\midrule
\midrule
Head Trauma & Wound / Normal & \textbf{2} if all match GT \\
\cmidrule{1-1}
Torso Trauma & & \textbf{1} if at least 2 match GT \\
\cmidrule{1-2}
Lower Ext. Trauma & Normal / Wound / & \textbf{0} otherwise \\
\cmidrule{1-1}
Upper Ext. Trauma & Amputation & \\
\midrule\midrule
Ocular Alertness     & Open / Closed / NT & \textbf{2} if all match GT \\
\cmidrule{1-2}
Verbal Alertness     & Normal / Absent /  & \textbf{1} if at least 2 match GT \\
\cmidrule{1-1}
Motor Alertness & Abnormal / NT & \textbf{0} otherwise \\
\bottomrule

\end{tabular}
\end{table}

System performance was evaluated using a structured scoring framework that awarded point scores for correctly identifying physiological signs and conditions under time constraints. Each casualty presented a set of target findings, and points were awarded when the system accurately reported those findings. In Table~\ref{scoring} we present a list of physiological measurements together with their corresponding score rubric. Additional weight was given for completing the assessment rapidly, within a ``golden time window" (GW), a period in which interventions are likely to prevent death, representing the urgency of real-world triage. The primary objective was to maximize the cumulative score for all casualties during the execution time budget of the scenario.

\begin{figure*}[htbp!]
    \centering
    \includegraphics[width=.8\linewidth]{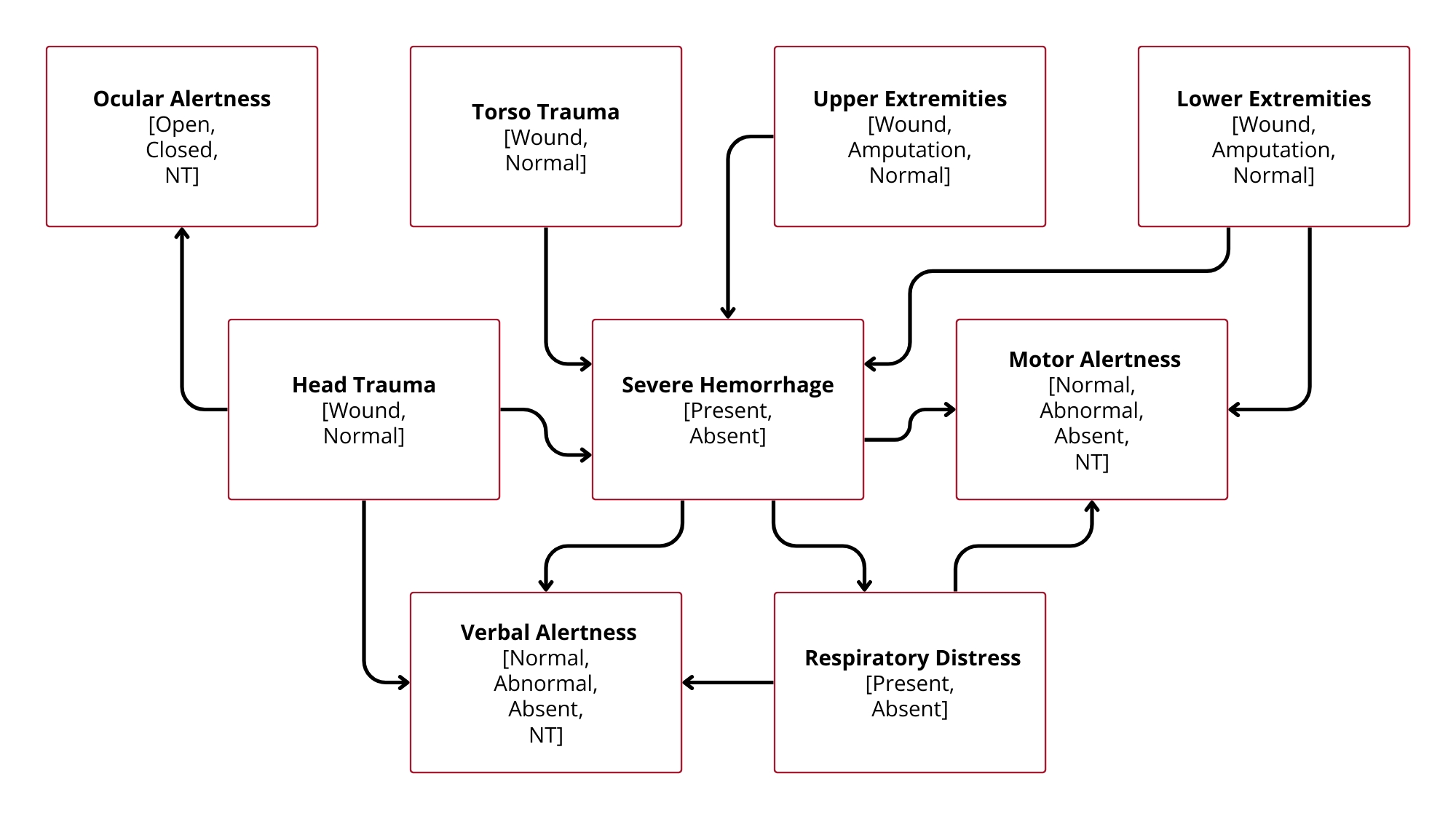}
    \caption{Bayesian network architecture illustrating the relationships among vital signs.}
    \label{fig:bn}
\end{figure*}

\subsection{System Architecture and Integration}

To address the multifaceted issues involved in the assessment of casualties, our solution leverages a probabilistic modeling framework grounded in domain expertise. The underlying BN model was constructed and processed using GeNIe Modeler~\cite{genie2022}. Its seamless integration into the Robot Operating System 2 (ROS 2)~\cite{maruyama2016exploring} stack was achieved through a custom interface developed with SMILE~\cite{smile2022}, which facilitates real-time inference and data exchange. This implementation follows the principles outlined in~\cite{druzdzel1999smile} for practical application within a modern robotic middleware ecosystem.

The perception system was deployed on a robotic platform equipped with a sensor suite comprising a high-resolution RGB camera, a microphone, a radar, a lidar and a thermal camera enabling the collection of visual and audio data from a stand-off distance. The system is built on the ROS 2 framework, where individual physiological estimators operate as independent nodes. Each node, corresponding to an input variable in the BN model shown in Figure~\ref{fig:bn}, is an independent AI model specialized in assessing a specific physiological sign from the robot sensor feed.

Each estimator node publishes time-stamped predictions (e.g., severe hemorrhage, ocular alertness) to the central BN service. When the robot signals scan completion or at a fixed cadence, the service triggers BN inference to impute unobserved variables and produces a triage posterior that is returned to the scoring module. The explicitly modular sensor/model-agnostic architecture integrates black-box estimator modules, exposing only their predicted variables, without any internal details. Components can therefore be swapped or added without modifying the reasoning layer. This modularity ensures that the system remains adaptable to evolving technologies in emergency response.

The BN treats incoming predictions as uncertain evidence and fuses heterogeneous sources with unknown or variable reliability. Upon receiving inputs, the BN performs the inference process detailed in Equation \ref{eq:pred} to estimate unobserved physiological variables and, through this probabilistic treatment, generates a consistent posterior over the patient state. This approach reduces the impact of errors in any single module, allows reasoning over incomplete or noisy inputs, and supports on-the-fly updates as new data become available, making it suitable for real-time triage.

The parameterization of the Bayesian Network, specifically the definition of the Conditional Probability Tables (CPTs) for each node, was carried out through a knowledge-driven engineering process. Due to the absence of publicly available standardized datasets for MCI scenarios, a data-driven approach was not feasible. Instead, the CPTs were populated based on qualitative rules and heuristics elicited through consultations with medical experts, as detailed in Section~\ref{subsec:3.3_expert}. This process involved translating qualitative statements (e.g., ``highly likely,'' ``possible,'' ``unlikely'') into quantitative probability values.

To ensure consistency and facilitate replicability, we adopted the following convention:
\begin{itemize}
    \item \textbf{Strong causal relationships} (e.g., a lower extremity amputation almost always causes severe hemorrhage) were modeled using probabilities in the range of 0.8 -- 0.95.
    \item \textbf{Moderate associations} (e.g., head trauma may be associated with abnormal verbal alertness) were assigned values in the 0.4 -- 0.6 range.
    \item \textbf{Weak dependencies} were represented by low probabilities, close to the prior probabilities of the baseline.
\end{itemize}

For instance, the probability for "Closed" for "Ocular Alertness" node in the case when prior of "Head Trauma" was "Wound" was set to 0.7. Although exact values were heuristically assigned, this structured methodology provides transparency and a framework for future model calibration should relevant data become available.

\subsection{Robotic Platform}

The experimental infrastructure is composed of a heterogeneous team of aerial and ground robots operating within a common ROS~2 framework. The robotic fleet consists of three Spot-type quadrupedal ground robots (left UGV in Figure \ref{fig:robots}), one custom-made aerial platform (top right UAV in Figure \ref{fig:robots}), and one Indago 4 Scout aerial platform (bottom right UAV in Figure \ref{fig:robots}). Both UAVs are equipped with GPS, a standard RGB camera, and a thermal camera. Their primary role is to rapidly detect, localize, and track casualties and to detect bleeding, leveraging their ability to cover large areas quickly and provide aerial situational awareness.

The Spot robots are equipped with a diverse set of sensors, including a LiDAR, GPS, multispectral camera, radar, microphone, depth camera, and thermal camera. Their task is twofold: first, to autonomously locate victims on the ground and second, to perform a detailed estimation of vital signs as defined in Table~\ref{scoring}. This multimodal sensing capability enables the collection of complementary information in detail that is not accessible from aerial platforms.  

To ensure consistent and reliable assessment, the predictions generated by each platform are processed through matching algorithms that aggregate observations corresponding to the same casualty. These fused data streams are then forwarded to the BN, which infers any missing information and produces a coherent triage profile. All communication and coordination between platforms are implemented using ROS 2 topics and services, ensuring modularity, scalability, and interoperability across the robotic system. For computations, our robots used NVIDIA 64 GB Jetson AGX Orin.

\subsection{Expert Knowledge Elicitation}
\label{subsec:3.3_expert}

The design and parameterization of the BN, whose structure is shown in Figure~\ref{fig:bn}, involved three domain experts, each with more than a decade of field experience in emergency medicine and rapid response medical interventions.

The methodology consisted of an iterative cycle. Initially, our engineering team conducted a series of individual interviews to identify the most critical observable indicators and their high-level relationships. Based on this qualitative input, an initial BN structure and preliminary Conditional Probability Tables (CPTs) were drafted. These drafts translated qualitative expert statements (e.g., "a lower extremity amputation almost always causes severe hemorrhage") into quantitative probability values.

In the second phase, the drafted model was presented back to the experts in follow-up sessions. Their role was to validate the logic of the model, identify potential gaps, and challenge assumptions. This feedback loop allowed the engineering team to manually refine the CPTs to ensure that the model's inferential behavior was balanced and consistent with the established clinical reasoning. This expert-in-the-loop approach, although not fully automated, ensured that the final model was computationally robust and grounded in real-world medical expertise.

\section{Results}
\label{sec:results}
detailed in Table~\ref{tab:performance_combined}.

The effectiveness of our framework was evaluated during the DTC Round 1 Systems Competition.
We report performance results for two scenarios. The third
run is excluded from this analysis due to a hardware-level failure prior to
deployment that prevented consistent data acquisition. Despite this, Team Chiron’s performance in
the remaining missions was sufficient to secure 4th place out of 11 teams overall.
Our findings demonstrate a significant improvement in the robot's triage capabilities when equipped with our Bayesian reasoning engine. 
The comparative analysis between BN-based cognitive architecture ("Robot + BN"), against a baseline system that relies solely on the direct outputs of its perception pipeline ("Robot") on 20 distinct casualty cases divided into two separate operationally realistic scenarios termed Run A and Run B are detailed in Table~\ref{tab:performance_combined}. 
Our integrated system more than doubled the baseline score in Run A (from 25 to 61) and nearly tripled it in Run B (from 16 to 45). This highlights the architecture's ability to overcome the limitations of real-world robotic perception, where sensor failures, environmental occlusions, and ambiguous visual cues are common.

\begin{table}[t]
\centering
\caption{Comparison of the system performance with and without BN}
\renewcommand{\arraystretch}{1} 
\label{tab:performance_combined}
\begin{tabular}{c c c c c}
\toprule
& \multicolumn{2}{c}{\textbf{Run A Scores}} & \multicolumn{2}{c}{\textbf{Run B Scores}} \\
\cmidrule(lr){2-3} \cmidrule(lr){4-5}
\textbf{Casualty Id} & \textbf{Robot} & \textbf{Robot + BN} & \textbf{Robot} & \textbf{Robot + BN} \\
\midrule\midrule
1  & 9 & 9 & 0 & 0 \\
2  & 0 & 3 & 0 & 3 \\
3  & 0 & 3 & 2 & 7 \\
4  & 0 & 3 & 2 & 7 \\
5  & 0 & 3 & 0 & 4 \\
6  & 0 & 5 & 1 & 3 \\
7  & 4 & 5 & 8 & 11 \\
8  & 0 & 7 & 1 & 3 \\
9  & 8 & 9 & 2 & 7 \\
10 & 4 & 7 & -- & -- \\
11 & 0 & 7 & -- & -- \\
\midrule\midrule
\textbf{Total}       & 25/132 & \textbf{61/132} & 16/108 & \textbf{45/108} \\
\bottomrule
\end{tabular}
\end{table}

\begin{table}[b]
\centering
\caption{Comparison of Correct Assignments and Assignment Attempts with and without BN across all casualties.}
\renewcommand{\arraystretch}{1} 
\label{tab:performance3}
\setlength{\tabcolsep}{5pt}
\begin{tabular}{l c c c c}
\toprule
& \multicolumn{2}{c}{\textbf{Correct Assignments}} & \multicolumn{2}{c}{\textbf{Assignment Attempts}} \\
\cmidrule(lr){2-3} \cmidrule(lr){4-5}
\textbf{Vital} & \textbf{Robot} & \textbf{Robot + BN} & \textbf{Robot} & \textbf{Robot + BN} \\
\midrule\midrule
Severe Hemorrhage     & 6 & 12 & 12 & 19 \\
Respiratory Distress  & 5 & 16 & 6 & 19 \\
Head Trauma           & 0 & 15  & 0 & 19 \\
Torso Trauma          & 0 & 11  & 0 & 19 \\
Lower Ext. Trauma     & 8 & 11 & 14 & 19 \\
Upper Ext. Trauma     & 3 & 8  & 14 & 19 \\
Ocular Alertness      & 0 & 8  & 1  & 19 \\
Motor Alertness       & 0 & 8  & 0  & 19 \\
Verbal Alertness      & 3 & 7  & 8  & 19 \\
\midrule\midrule
\textbf{Total}        & 25 & 96 & 55 & 171 \\
\cmidrule(lr){2-3}\cmidrule(lr){4-5}
\textbf{Reliability} & -- & -- & 0.31 & \textbf{0.95} \\
\cmidrule(lr){2-3}\cmidrule(lr){4-5}
\textbf{Performance} & 14\% & \textbf{53\%} & -- & -- \\
\cmidrule(lr){2-3}\cmidrule(lr){4-5}
\textbf{Accuracy} & 46\% & \textbf{56\%} & -- & -- \\
\bottomrule
\end{tabular}
\end{table}

Table~\ref{tab:performance3} further details this improvement across three key metrics. System's \textbf{Reliability}, which measures the robots' ability to provide an assessment even with partial data, increased from 0.31 to 0.95. This indicates that the baseline failed to make an assessment in 69\% of cases, whereas our integrated system provided an output in 95\% of them. Achieving 100\% reliability was not achievable due to the inability of the system to locate one of the casualties in Run B. Furthermore, the quality of these assessments improved, with \textbf{Accuracy} (correct assignments among those attempted) rising from 46\% to 56\%. 

Most importantly, the overall \textbf{Performance}, which measures the rate of correct assignments across all possible vitals, increased nearly four-fold from 14\% to 53\%. This was largely attributable to the BN's ability to infer missing information and resolve conflicting perceptual cues, resulting in a 3.84-fold increase in the total number of correctly assigned vitals. This underscores the critical role of the cognitive architecture in transforming the fragmented, unreliable outputs of a standard robotic perception pipeline into a coherent and actionable assessment, a crucial step for deploying autonomous systems in high-stakes scenarios.

\begin{figure}[h]
    \centering
    \includegraphics[width=.8\linewidth]{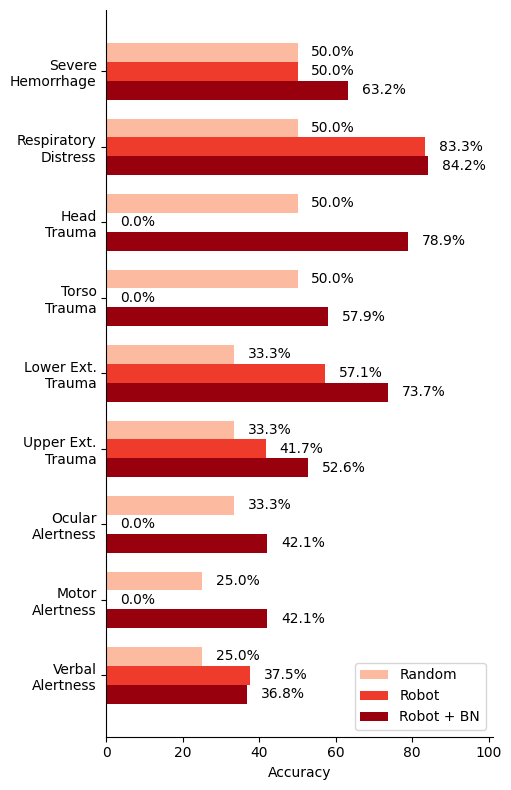}
    \caption{Comparision of random classifier, robot and robot + BN}
    \label{fig:robots_res}
\end{figure}

Figure~\ref{fig:robots_res} presents a comparative analysis of performance accuracy, distinguishing between configurations with and without the integration of the BN, relative to a random classifier baseline that uniformly selects from the set of valid labels. The inclusion of the BN demonstrably enhanced the system's performance, enabling it to surpass the random classifier across all evaluated vital signs. 

\subsection{Real-time Performance}

One design objective was to ensure our solution could operate in real time on field-deployable hardware. Our implementation achieves inference times of less than \textbf{1 millisecond} per update and maintains a memory footprint below \textbf{100 megabytes}. This performance enables deployment on resource-constrained platforms, such as the NVIDIA Jetson series or even a Raspberry Pi, without compromising the robot's responsiveness or reliability in real-world scenarios.

\section{Discussion}
\label{sec:discussion}

Our results demonstrate that integrating a Bayesian reasoning engine with advanced vision-based sensing can produce a substantial improvement in the casualty assessment capabilities of an autonomous robotic system operating in the field. This highlights the profound impact of explicitly modeling uncertainty and intermodal dependencies, a capability that standard robotic perception pipelines lack. The system's performance improves as it gathers more evidence, emulating the iterative reasoning of human responders at machine speeds. This shift from static, one-off assessments to continuous probabilistic reasoning is a key step towards building truly effective and useful autonomous triage robots.

This research underscores a critical limitation of many current robotic systems: reliance on isolated perception modules, which often function as "black boxes" in isolation from domain knowledge. In the high-stakes, chaotic environment of an MCI, this can lead to inconsistent or medically implausible predictions. For example, a robot's perception pipeline might correctly identify a limb amputation while another module, seeing minor movements, incorrectly reports normal motor alertness. Our cognitive architecture, centered on the Bayesian Network, resolves such contradictions by enforcing clinical constraints. This offers a key advantage in system-level robustness and is paramount for any robot deployed in a safety-critical role.

A fundamental requirement for any field-deployable robot is fault tolerance. Our architecture enables graceful degradation when perceptual data is incomplete. A failure of a single input module, a common occurrence due to sensor occlusions, communication loss, or algorithmic errors, does not cause a catastrophic failure of the overall case assessment task. Instead, the BN marginalizes over the missing variable, providing the best possible assessment based on the available data from all other functioning sensors. This operational reliability is essential for autonomous systems operating outside controlled lab environments.

Furthermore, the probabilistic framework is inherently extensible, making the system's cognitive architecture future-proof. Although the current implementation treats categorical inputs as direct evidence, the BN structure readily accommodates future enhancements. As new sensors or improved perception algorithms are added to the system, their outputs can be integrated with minimal changes to the core reasoning engine. It could also be extended to incorporate model confidence scores or to implement mechanisms for rejecting spurious sensor readings, further enhancing robustness. This forward-looking flexibility was a key factor in our architectural design.

Although our initial results are encouraging, a more comprehensive evaluation is in order. Future work should involve direct comparisons with existing deterministic and probabilistic approaches, including rule-based expert systems, hybrid neuro-symbolic models, and hybrid deep learning pipelines.

\section{Conclusion and future work}
\label{sec:conclusion}

In this work, we introduced a cognitive architecture to address the critical challenge of robust decision making for autonomous robots deployed in mass casualty incidents. By centering our system on an expert-guided Bayesian Network, we enabled it to fuse fragmented and uncertain data from its perception pipeline into a coherent multifactor and medically plausible triage assessment. The field evaluations conducted during the DTC Round 1 demonstrated that this architecture can substantially improve reliability and overall performance, transforming the presented system from a simple data collector into an effective decision-making agent. This work represents a key step towards deploying autonomous systems in safety-critical real-world scenarios.

Based on our findings, we propose three primary directions for future research to build upon the foundation presented.

\subsection{Human-Robot Teaming and Ethical Deployment}
In any realistic deployment, the robots will operate as part of a team with human responders. Future research must focus on the principles of effective human-robot teaming. This includes designing intuitive interfaces for communicating the system-generated assessments and corresponding uncertainties to a human operator, enabling shared situational awareness and collaborative decision making. As with any autonomous system operating in high-stakes contexts, ethical considerations are paramount. Future development must incorporate robust safeguards and human-in-the-loop oversight mechanisms to ensure that autonomous triage systems are deployed responsibly and effectively as tools to aid, not replace, human experts.

\subsection{Robustness Testing}
Our experiments have demonstrated a significant improvement in key metrics, confirming the effectiveness of the presented approach. However, a fundamental conclusion drawn from our analysis is the critical importance of robustness testing. This is especially critical given the model's multi-modal architecture, which relies on inputs generated by other models. Such structures are susceptible to cascading error propagation, where even minor inaccuracies in one of the input modules could lead to significant errors in the final output, even though we have not observed such effects in our experiments.

\subsection{Data Quality and Availability.}
A fundamental challenge in this work was the complete absence of empirical data that could be used for machine learning. In our application domain, there are no relevant publicly available standardized datasets. In addition, the specific nature of the triage task imposes extremely high data quality requirements. Any error in the classification process can lead to severe consequences and untoward patient outcomes; therefore, any training dataset would need to be free from annotation errors. We are pushing to collect relevant data during our field experiments, but we would like to encourage the research community to join us in these efforts. 

\section*{ACKNOWLEDGMENT}
This work has been partially supported by the Defense Advance Research Projects Agency (award
HR00112420329) and National Science Foundation (awards 2427948 and 2406231). The views,
opinions, and/or findings expressed are those of the authors and should not be interpreted as representing the official views or policies of the Department of Defense or the U.S. Government. We also
thank the DARPA Triage Challenge organizers and the technical observers for their logistical support
during the field evaluations
\bibliographystyle{IEEEtran}
\bibliography{main}

\end{document}